\documentclass{article}
\usepackage{spconf,amsmath,graphicx}
\usepackage{hyperref}
\usepackage{multirow}
\usepackage{algorithm}
\usepackage{algpseudocode}
\usepackage{amsmath}
\usepackage{graphics}
\usepackage{epsfig}

\algnewcommand\algorithmicforeach{\textbf{for each}}
\algdef{S}[FOR]{ForEach}[1]{\algorithmicforeach\ #1\ \algorithmicdo}

\title{SMILE: Sequence-to-Sequence Domain Adaptation\\with Minimizing Latent Entropy for Text Image Recognition}
\name{Yen-Cheng Chang${^1}$\qquad
Yi-Chang Chen${^1}$\qquad
Yu-Chuan Chang${^1}$\qquad
Yi-Ren Yeh${^2}$}
\address{$^1$E.SUN Financial Holding CO., LTD., Taiwan\\
$^2$Department of Mathematics, National Kaohsiung Normal University, Taiwan\\ \\
{\tt\small \{ycchang-21549, ycchen-20839, steven-20841\}@esunbank.com.tw} \\
{\tt\small yryeh@nknu.edu.tw}
}

\begin{document}
%

\maketitle
\begin{abstract}
Training recognition models with synthetic images have achieved remarkable results in text recognition. However, recognizing text from real-world images still faces challenges due to the domain shift between synthetic and real-world text images. One of the strategies to eliminate the domain difference without manual annotation is unsupervised domain adaptation (UDA). Due to the characteristic of sequential labeling tasks, most popular UDA methods cannot be directly applied to text recognition. To tackle this problem, we proposed a UDA method with minimizing latent entropy on sequence-to-sequence attention-based models with class-balanced self-paced learning. Our experiments show that our proposed framework achieves better recognition results than the existing methods on most UDA text recognition benchmarks. All codes are publicly available\footnote{\url{https://github.com/timtimchang/SMILE}}.


\end{abstract}


\section{Introduction}
\label{sec:introduction}



Deep learning methods are widely used in text recognition tasks \cite{crnn, what-wrong}. It often requires enormous manual labeling data to train deep learning models with large parameters. To reduce the burden of manual annotation, generating artificial text images by synthetic image engines has achieved remarkable results in text recognition \cite{esun-synth, mjsynth, synthtext}. However, recognizing text from real-world images still faces challenges due to the domain shift caused by the large variance in text styles, viewpoints, and backgrounds between synthetic and real-world text images.

\begin{figure}[t]
\centerline{\includegraphics[width=8cm]{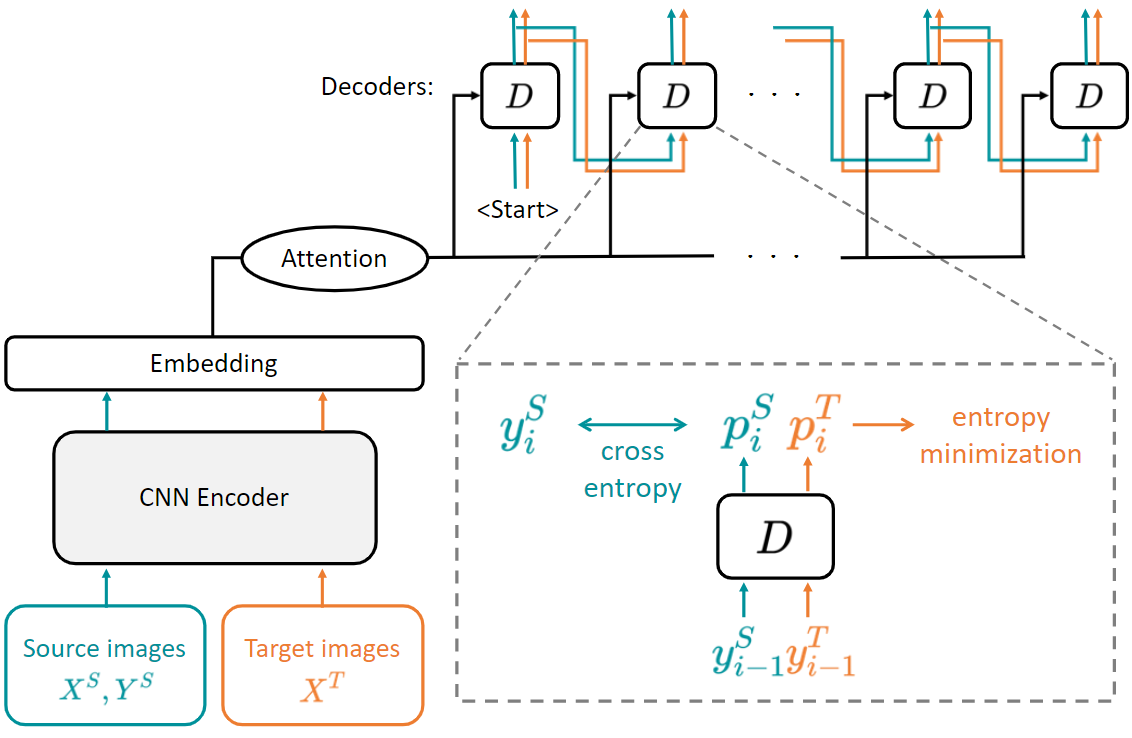}}
\caption{The network structure of our SMILE. $D$ is the decoder for the sequential outputs. $p^{S}$ and $p^{T}$ are the sequential output probabilities of labeled source and unlabeled target text images.}
\label{fig1}
\end{figure}


The most common strategy to eliminate the domain difference is using the pre-trained model from source domain data and fine-tuning the model with target domain data where manual annotating is still required. On the other hand, many researchers have proposed unsupervised domain adaptation (UDA) to couple the labeled source domain and unlabeled target domain to learn a domain-invariant model \cite{Gretton et al. 2007,Baochen Sun and Kate Saenko. 2016, JunbaoZhuo 2017,Yaroslav Ganin and Victor Lempitsky. 2015, Zhongyi Pei et al. 2018, EricTzeng et al. 2017,F. Zhan et al. 2019, Zhang2021}. Our work aims to apply UDA with labeled synthetic text images and unlabeled real-world text images to learn an improved text recognition model in the target domain (i.e., real-world text images) by comparing the model learned from synthetic text images. However, most popular UDA methods cannot be directly applied to sequential labeling tasks, such as text recognition. More specifically, those UDA methods only work on a global representation and did not consider the sequential outputs with local character-level information in text recognition. 

To tackle the sequential labeling for domain adaptation, we proposed an unsupervised \textbf{S}equence-to-sequence domain adaptation with \textbf{M}inim\textbf{I}zing \textbf{L}atent \textbf{E}ntropy (SMILE) based on the attention-based model as shown in Figure \ref{fig1}. There are two crucial aspects in UDA: ensuring the validity of the source domain and reducing the discrepancy between the source domain and target domain. Our SMILE adopts conventional cross-entropy loss for labeled source domain data as the first aspect. For the second aspect, inspired by \cite{Grandvalet2005}, SMILE applies entropy minimization to sharpen the prediction probability of each unlabeled latent representation of target domain data from its decoders in the sequence-to-sequence attention-based model. We minimize each latent representation's classification loss and discrepancy loss for source and target domains during the training process, respectively. To avoid selecting noisy pseudo labels from unlabeled data, we also apply class-balanced self-paced learning in our proposed framework to improve performance. More details will be addressed in Section \ref{sec:methodology}.

\section{Related Work}
\label{sec:related_work}

To reduce manual annotation efforts, many researchers have proposed unsupervised domain adaptation to learn a shared space between the unlabeled target domain and the source domain data, such as Maximum Mean Discrepancy  (MMD) \cite{Gretton et al. 2007}, correlation alignment distance (CORAL) \cite{Baochen Sun and Kate Saenko. 2016, JunbaoZhuo 2017}, adversarial loss \cite{Yaroslav Ganin and Victor Lempitsky. 2015, Zhongyi Pei et al. 2018, EricTzeng et al. 2017,Kang et al. 2020}. However, most of these existing methods optimize the global representation by minimizing the aforementioned domain shift measures. For example, \cite{Kang et al. 2020} uses a temporal pooling to aggregate the information of sequences (i.e., character-level) for a discriminator network in handwritten word recognition. GA-DAN generates real scene text images by converting synthetic images but omits domain shift in fine-grained character-level images. Unlike minimizing the domain difference only with the global representation for sequence-to-sequence tasks, an adversarial sequence-to-sequence domain adaptation (ASSDA) \cite{Zhang2021} is proposed to address the issue by the sequential architecture. The authors use global and local alignment to minimize the domain difference to address fine-grained character level domain shift. Different to ASSDA, we design a discrepancy loss to minimize the entropy of output probability for each unlabeled latent representation of the target domain from its decoders.

\section{Methodology}
\label{sec:methodology}



In our work, we proposed an unsupervised domain adaptation for text recognition. More specifically, we have labeled synthetic text images and unlabeled real-world text images as the source and target domain data, respectively. Let $X^{S} = \{ \mathbf{x}_{1}^{S}, \mathbf{x}_{2}^{S}, ..., \mathbf{x}^{S}_{|N^{S}|} \}$ and $X^{T} = \{ \mathbf{x}_{1}^{T}, \mathbf{x}_{2}^{T}, ..., \mathbf{x}^{T}_{|N^{T}|} \}$ represent the source and target domain instances where $N^{S}$ and $N^{T}$ are the corresponding data sizes. The ground truth of each instance from the source domain is defined as  $\mathbf{y}^{S} = \{ y_{1}^{S}, y_{2}^{S}, ..., y_{L}^{S}\} \in Y^{S}$, where $L$ is the variable length of the text sequence. We aim to learn an improved text recognition model in the target domain (i.e., unlabeled real-text images).

\subsection{Text Recognition with Attention-base Model}
\label{ssec:ocr_framework}


The framework of our proposed method is based on the sequence-to-sequence attention-based model in \cite{what-wrong}. The text recognition task can be considered as an encoding procedure between input text images $\mathbf{x} \in X$ to a sequence-like text groud truth where $\mathbf{y} \in Y$. As shown in Figure \ref{fig1}, the decoder ($\mathbf{D}$) will generate sequential predicted labels. In our framework, 
the cross-entropy loss of each predicted label in the sequence is minimized to learn the recognition ability from source domain data as follows:


\begin{equation}
\label{eq1}
 L_{dec} = E_{\mathbf{x}^{S}\in X^{S},\mathbf{y}^{S} \in Y^{S}} \{ -\sum^{L}_{i}log \ p(y^{S}_i| \mathbf{x}^{S}) \}.
\end{equation}
However, the decoder loss $L_{dec}$ only guarantees to learn the recognition ability for labeled synthetic images. It cannot be fully transferred to unlabeled real-world images of interest. To tackle this problem, we make use of entropy minimization to eliminate the domain difference between labeled synthetic images and real-world images. More detail will be illustrated in the following subsection.



\subsection{Sequence-to-sequence UDA with Minimizing Latent Entropy}
\label{ssec:entropy_minimization}

Entropy minimization has been widely used in semi-supervised learning and unsupervised domain adaptation \cite{Grandvalet2005, SaitoICCV2019, WangPR2021}. The goal is to enforce the unlabeled examples with slight overlapping between classes for reducing the discrepancy between source and target domain. However, entropy minimization cannot directly be applied to text recognition tasks due to the sequential labels in a text image. As shown in Figure 1, the sequence-to-sequence attention-based model generates sequential character-level predictions for unlabeled target domain data. To achieve entropy minimization for sequential labeling problems, we minimize the individual entropy from the character-level prediction within the sequential output. More specifically, we have a single entropy value to represent the respective distribution for each predicted character in the sequence. The entropy minimization loss of a single predicted sequence is the summation of all the entropy from each predicted character in a sequence instance. The entropy minimization loss of the unlabeled target domain is defined as follows:

\begin{equation}
\label{eq2}
L_{ent} = E_{\mathbf{x}^{T}\in \mathbf{X}_{T}} \{ \sum_{i}^{L} -log \  p(y^{T}_i | \mathbf{x}^{T}) \}.
\end{equation}
The entropy minimization loss optimizes the prediction of the target domain without supervision. 

In our proposed UDA for text recognition, we integrate the objective functions \eqref{eq1} and \eqref{eq2} to sharpen the predicted class distribution from the target domain while the recognition ability from the source domain is maintained. The integrated loss is defined as follows:

\begin{equation}
\label{eq3}
L_{SMILE} = L_{dec} + \lambda L_{ent},
\end{equation}
where $\lambda$ is the weight of the entropy minimization loss. The parameter $\lambda$ controls the trade-off of the losses between source and target domains for the optimization, and we simplely use  $\lambda = 1$ in our experiments.

\begin{table*}[t]
{\small 
\begin{center}
\begin{tabular}{lcc|c|lclc|lc}
\hline  
\multicolumn{3}{c|}{    } & &\multicolumn{4}{c|}{\text { \textbf{(a) regular}   } } & \multicolumn{2}{c}{\text { \textbf{(b) irregular} } } \\
\text { \textbf{Model} } & \text { \textbf{Year} } & \text { \textbf{UDA} }  & \text { \textbf{Train data} } & \text { \textbf{IIIT5K} } & \text { \textbf{SVT} } & \text { \textbf{IC03} } & \text { \textbf{IC13} } & \text { \textbf{IC15} } & \text { \textbf{CUTE} } \\
\multicolumn{3}{c|}{    } & & \text { 3000 } & \text { 647 } & \text { 860 } & \text { 857 }  & \text { 1811 } & \text { 288 }  \\

\hline
CRNN \cite{crnn, what-wrong}       &  2015  & N    & MJ & 82.9          & 81.6          & 93.1          & 91.1          & -             & -    \\
RARE \cite{rare, what-wrong}   &   2016   & N    & MJ & 86.2          & 85.8          & 93.9          & 92.6          & 74.5          & 70.4 \\
STAR-Net \cite{starnet, what-wrong}    &  2016  & N    & MJ & 87.0          & 86.9          & 94.4          & 92.8          & 76.1          & 71.7 \\
GRCNN \cite{grcnn,what-wrong}      &  2017  & N    & MJ+PRI & 84.2          & 83.7          & 93.5          & 90.9          & -             & -    \\
Char-Net \cite{charnet,what-wrong}   & 2018   & N    & MJ & 83.6          & 84.4          & 91.5          & 90.8          & 60.0          & -    \\
STR2019 \cite{what-wrong}       &2019 & N    & MJ+ST & \textbf{87.9} & \textbf{87.5} & 94.9          & \textbf{93.6} & 77.6          & 74.0   \\
SSDAN-base \cite{ssdan}  &  2019  & N    & MJ+ST & 81.1          & 82.1          & 91.2          & 91.0          & -             & -    \\
SMILE-base / ASSDA-base \cite{Zhang2021} &  & N    & MJ+ST & 87.5          & 86.7          & \textbf{95.1}          & 92.9          & \textbf{78.1}          & \textbf{74.2} \\
\hline
SSDAN \cite{ssdan} &    2019     & Y    & MJ+ST & 83.8          & 84.5          & 92.1          & 91.8          & -             & -             \\
ASSDA \cite{Zhang2021}  &    2021    & Y    & MJ+ST & 87.6          & \textbf{87.8} & 95.5          & 93.8          & 78.7          & \textbf{76.3} \\
SMILE (w/ self-paced)        &  & Y    & MJ+ST & \textbf{89.3} & 87.6          & \textbf{96.0} & \textbf{94.9} & \textbf{78.9} & 75.6          \\
\hline
SMILE-finetune / ASSDA-finetune \cite{Zhang2021} &   &  & MJ+ST & 89.7 & 87.3          & 94.4          & 94.3          & 79.7          & 74.9 \\
\hline
\end{tabular}
\caption{Results of our SMILE on both regular and irregular datasets compared with existing methods. UDA methods  are all pre-trained with synthetic MJ \cite{mjsynth} and ST \cite{synthtext} datasets. SMILE/ASSDA-base represents only training the recognition model with labeled source data. SMILE/ASSDA-finetune represents pre-training and fine-tuning the recognition model with labeled source and target text images. Note that the pre-trained model used in SMILE and ASSDA are the same.}
\label{table:domain_adpation_result}
\end{center}
}
\end{table*}

{\small 
\begin{algorithm}[t]
  \caption{SMILE with class-wise self-paced learning}
  \label{alg::SMILE}
  \begin{algorithmic}[1]
    \Require
      $X^{S}$: source domain images;
      $Y^{S}$: source domain labels;
      $X^{T}$: target domain images;
      $\mathcal{P}_{init}$: initial target portion;
      $\mathcal{P}_{add}$: adding target portion;
      $\lambda$: weight
    \For{t = 1 to T}
      \State $\{p_i^{S}\}$: probability outputs of $X^{S}$ from decoder $D$
      \State $\{p_i^{T}\}$: probability outputs of $X^{T}$ from decoder $D$
      \State $\hat{C}$: pseudo labels determined by $\{p_i^{T}\}$
      \State $ChoosedLosses:null\ array$
      \ForEach {class $c \in \hat{C}$}
        \State  $\{p_i^{c}\}$ : probability outputs in $\{p_i^{T}\}$ for class c
        \State $EntList_{c} = \{ Entropy(p), \forall p \in \{p_i^{c}\} \}$
        \State $\mathcal{P}_{t} = \min (\mathcal{P}_{init} + \mathcal{P}_{add} * t, 1)$
        \State $k_{c} = \lceil |\{p_i^{c}\}| * \mathcal{P}_{t}  \rceil $
        \State $EntList^{topk}_{c} = topk(EntPool_{c}, k=k_{c})$
        \State $ChoosedLosses.extend(EntList^{topk}_{c})$
      \EndFor
    \State $L_{dec} = CrossEntropy ( \{p_i^{S}\}, Y^{S} )$ 
    \State $L_{ent} = Mean ( ChoosedLosses )$ 
    \State $L_{SMILE} = L_{dec} + \lambda L_{ent}$
    \EndFor
  \end{algorithmic}
\end{algorithm}
}

\subsection{Class-balanced Self-paced Learning}
\label{ssec:class-balanced_self-paced_learning}

In our proposed framework, calculating the entropy minimization loss of entire target domina data at the beginning of the training might fail to achieve better domain adaptation due to the unstable prediction quality. In \cite{Zou2018}, the authors proposed self-paced learning to follow the scheme of "easy-to-hard" by a curriculum learning order, where one selects the most confident predictions from target domain instances into the training pool and gradually increases the portion of target instances. 

However, an imbalanced problem is common in text recognition, especially for special text symbols or multilingual scenarios. To tackle this problem, we apply the class-balanced self-paced learning strategy \cite{Zou2018} to increase stability during the training procedure. More specifically, we assume that a character-level prediction with lower entropy is more confident. The training process can start with more confident character-level predictions, and we gradually select these more confident predictions. In Algorithm 1, $\mathcal{P}_{init}$ decides the initial portion of each class in the target domain that we expect to select for entropy minimization at the initial step. On the other hand, $\mathcal{P}_{add}$ is the portion of target domain data selected for training at the following steps. The detailed process of the class-balanced self-paced learning with latent entropy minimization is presented in Algorithm \ref{alg::SMILE}.



\section{Experiments}
\label{sec:experiments}
To compare our proposed method with existing methods, our experiments' training and evaluation procedures follow the same protocol in \cite{what-wrong, Zhang2021}. In the protocol, we pre-trained the model with synthetic scene text datasets as the source domain and validated on real scene text datasets as the target domain. There are two types of real scene text datasets for the evaluation in our experiments: regular scene text and irregular scene text. The detail of the implementation and experiment will be illustrated in the following subsections.

\subsection{Implementation Details}
\label{ssec:experiments_settings}

Our experiments follow the TPS-ResNet-BiLSTM-Attn framework in \cite{what-wrong} for training a recognition model. The number of fiducial points of TPS, the number of output channels of ResNet, and the size of the BiLSTM hidden state are 20, 512, and 256, respectively. The setting of optimizer used in our experiments is also the same in \cite{what-wrong, Zhang2021} for fair comparisons.


We combine MJSynth (8.9M) \cite{mjsynth} and SynthText (5.5 M) \cite{synthtext} as our synthetic scene text datasets in the source domain in our experiments. The texts are rendered onto real-wold images, and all of these text images are well annotated. In addition to synthetic datasets, we also have two types of real-world text images in the target domain: regular and irregular texts. The irregular text could have harder curved, rotated, or distorted instances than the regular text.

\subsection{From Synthetic to Regular Text Images}
\label{sssec:domain_adaption_on_regular}


Our proposed UDA method is applied from synthetic images to regular real-world text images in our first experiment. We consider four real-world scene text recognition datasets: IIIT5K-Words (IIIT), Street View Text (SVT), ICDAR2003 (IC03), ICDAR2013 (IC13). The details of the data description, such as data size and protocol, could be found in \cite{what-wrong, Zhang2021}.



The test results of real scene text datasets are shown in Table~\ref{table:domain_adpation_result}(a). Our experiments firstly consider the baseline model, SMILE-base, only trained with synthetic images to validate our proposed method. Compared to the baseline model, our SMILE achieves improved performance in all the regular text datasets. It shows that our SMILE can transfer the recognition ability from the labeled source domain to the unlabeled target domain. Our SMILE also outperforms SSDAN and ASSDA, which are the state-of-the-art methods in UDA for text recognition. Furthermore, our SMILE can achieve comparable results compared to the fine-tuned model, SMILE-finetune, which is fine-tuned with labeled target domain data. The results show that our proposed method could utilize the information from unlabeled target domain data.

\subsection{From Synthetic to Real Irregular Text Images}
\label{sssec:domain_adaption_on_irregular}


In our second experiment, we use ICDAR2015 (IC15) and CUTE80 (CUTE) as our irregular evaluation datasets, which could contain harder curved, perspective, or distorted instances. The details of the data description can also be found in \cite{what-wrong, Zhang2021}. 

As shown in Table~\ref{table:domain_adpation_result}(b), our SMILE consistently improves the performance compared to the baseline model by taking the unlabeled target data into account. Besides, our SMILE could produce comparable results compared to the state-of-the-art methods and fine-tuned model. In CUTE dataset, our SMILE is outperformed by ASSDA. The possible reason could be that the target domain data is relatively less and more diverse (carved), and our class-wise self-paced learning from easy to hard could over-fit those easier characters.

\subsection{Analysis of Self-paced Learning}
\label{ssec:analysis_of_self-paced_learnin}




This experiment discusses the influence of the parameters for the initial portion $\mathcal{P}_{init}$ and the adding portion $\mathcal{P}_{add}$ in our class-balanced self-paced learning. We consider different combinations of $\mathcal{P}_{init} $ = [0, 0.3, 0.5] and $\mathcal{P}_{add}$ = [$1e-4$, $5e-5$] for the class-balanced self-paced learning. The results of these combinations are presented in Table~\ref{table:self_paced_result}, and show that training the models with self-paced learning achieves better than without self-paced learning (i.e., ($\mathcal{P}_{init}$,$\mathcal{P}_{add}$) = (1,0)). It shows that our class-balanced self-paced learning could prevent the training model from selecting excessive noisy pseudo-labeled target data at the early stage of training.

Besides, the results also show that smaller $\mathcal{P}_{init}$ and $\mathcal{P}_{add}$ are preferred since it is expected to select more confident predictions at each step. As shown in Table~\ref{table:self_paced_result}, $\mathcal{P}_{init}=0.0$ and $\mathcal{P}_{add}=5e-5$ achieves best results. We use this combination to compare the existing method in Table~\ref{table:domain_adpation_result}.

\begin{table}[t]
{\footnotesize 
\begin{center}
\begin{tabular}{l|lclc|lc}
\hline  
\multicolumn{1}{c|}{ } & \multicolumn{4}{c|}{\text { \textbf{regular}   } } & \multicolumn{2}{c}{\text { \textbf{irregular} } } \\
$\mathbf{(\mathcal{P}_{init}, \mathcal{P}_{add})}$  & \textbf{IIIT5K} & \textbf{SVT} & \textbf{IC03}  & \textbf{IC13} & \textbf{IC15} &  \textbf{CUTE} \\
\hline
(0.0, 1e-4)  & 89.2          & 87.5          & \textbf{96.0} & \textbf{94.9}        & 78.0          & 74.6                 \\
(0.3, 1e-4)                     & 89.3          & 87.6          & 95.7  & 94.6      & 77.7          & 74.6          \\
(0.5, 1e-4)                        & \textbf{89.5} & 87.2          & 95.6           & 94.5          & 77.7 & 74.6           \\
(0.0, 5e-5)$^*$ & 89.3          & 87.6          & \textbf{96.0} & \textbf{94.9}     & \textbf{78.9} & \textbf{75.6}          \\
(0.3, 5e-5)                          & 89.2          & \textbf{88.1} & \textbf{96.0}& 94.7          & 77.7 & 74.6           \\
(0.5, 5e-5)                         & 88.5          & 87.6          & 95.3          & 94.2          & 78.0& 73.9            \\
(1.0, 0.0)    & 88.6          & 86.9          & 95.3          & 94.2                & 78.0          & 73.9                  \\
\hline
\multicolumn{5}{c}{$^*$ represents the parameters used in our experiments}&&
\end{tabular}
\caption{Parameter sensitivity of the parameters $\mathcal{P}_{init}$ and $\mathcal{P}_{add}$ in the class-balanced self-paced learning.}
\label{table:self_paced_result}
\end{center}
}
\end{table}

\section{Conclusion}
\label{sec:conclusion}




This paper proposed a sequence-to-sequence unsupervised domain adaptation with latent entropy minimization (SMILE) for text recognition. To make entropy minimization adaptive to the sequential labeling problem, we dismantle the sequential outputs in our proposed method to minimize the individual entropy with these character-level predictions. Besides, we also apply class-balanced self-paced learning to gradually select target domain data with noisy pseudo labels for maintaining stability during the model training. Our experiments show that our SMILE transfers the recognition ability from the labeled source domain to the unlabeled target domain and outperforms the state-of-the-art methods in most real scene text datasets.

\bibliographystyle{IEEEbib}
\bibliography{strings,refs}

\end{document}